\documentclass[twoside,a4paper]{llncs}
\usepackage{graphicx}
\usepackage{subcaption}
\usepackage{caption}
\usepackage{hhline}
\usepackage{siunitx}
\usepackage{amsmath}
\usepackage{multirow}
\usepackage{blindtext, xcolor}
\usepackage[draft]{todonotes}   
\newcommand{\vR}[1]{{#1}}  
 
\begin{document}
\mainmatter
\title{Learn the new, keep the old: Extending pretrained models with new anatomy and images} 
\author{\vR{Firat Ozdemir$^*$ \and Philipp Fuernstahl$^\dagger$ \and Orcun Goksel$^*$}}
\institute{\vR{$^*$Computer-assisted Applications in Medicine, ETH Zurich, Switzerland\\
$^\dagger$CARD Group, University Hospital Balgrist, University of Zurich, Switzerland}}

\maketitle

\begin{abstract}
Deep learning has been widely accepted as a promising solution for medical image segmentation, 
given a sufficiently large representative dataset of images with corresponding annotations.
With ever increasing amounts of annotated medical datasets, it is infeasible to train a learning method always with all data from scratch.
This is also doomed to hit computational limits, e.g., memory or runtime feasible for training.
Incremental learning can be a potential solution, where new information (images or anatomy) is introduced iteratively.
Nevertheless, for the preservation of the collective information, it is essential to keep some ``important'' (i.e.,\ representative) images and annotations from the past, while adding new information.
In this paper, we introduce a framework for applying incremental learning for segmentation and propose novel methods for selecting representative data therein.
We comparatively evaluate our methods in different scenarios using MR images and validate the increased learning capacity with using our methods.

\end{abstract}

\section{Introduction}
\label{sec:introduction}

With the growing interest in automatic and semi-automatic analysis of patients, available data size for research is continuously increasing.
Even for a single anatomical structure, soon it may become infeasible to retrain a network when a newly available data is introduced. 
On the other hand, one can expect to see variations in image properties across iterations of new data due to various factors, e.g.\ mechanical differences across imaging device brands, physiological differences across imaged subjects.
Furthermore, although various datasets from similar modality are often available, they belong to different studies, hence they have different field-of-view (FOV), image acquisition parameters, and/or annotated anatomy.
For instance, some MR modalities (i.e.,\ ultra short TE) are often used to analyze bones and tendons thanks to its high contrast. 
However, a study on diagnosis or healing quantification of Achilles tendon often do not allocate resources for proximal bone tissue annotation.
Similarly, for an osteotomy planning, often only bones are annotated to generate surgical guides. 
Aggregation of annotation knowledge across different anatomy, modality \& dataset are not well investigated; however, it is of growing interest in the machine learning community~\cite{hinton2015distilling,li2016learning} as in the form of an ``evolving'' classifier.
To the best of our knowledge, increasing label problem has neither been tackled for segmentation nor in medical community.

For class-incremental learning problem, initial works include finetuning~\cite{girshick2013rich}; however, it is well known~\cite{MCCLOSKEY1989catastrophic} that this results with ``catastrophic forgetting.'' 
Later on, learning without forgetting (LwF)~\cite{li2016learning} has been proposed, which utilizes distillation loss~\cite{hinton2015distilling} such that when new classes are being added to a network, final activation response of the previous classes are also used for backpropagation.
With iCaRL~\cite{rebuffi2016icarl}, authors extend on LwF by proposing a strategy for selecting an \textit{exemplar} dataset, which keeps a ``representative'' subset of the earlier training data for the existing classes, and put an upper bound on required memory requirements.
In~\cite{yang2017suggestive}, authors suggest a novel way to pick representative samples to train on, for the purpose of maximizing performance for binary segmentation task of gland cells at a next training iteration. 

In this work, our novelties are in line with blocks that are necessary to expand class-incremental learning for segmentation task; \textit{extending distillation loss to segmentation} without an assumption on mutual exclusivity of classes. 
We propose alternative methods for picking representative samples to sustain segmentation accuracy of prior classes.
To the best of our knowledge, this is the first work extending distillation loss for class-incremental segmentation.

\section{Incremental Head Networks}
\label{sec:methods}

\begin{figure}[!t]
    \centering
    \includegraphics[
            width = \textwidth
            ]{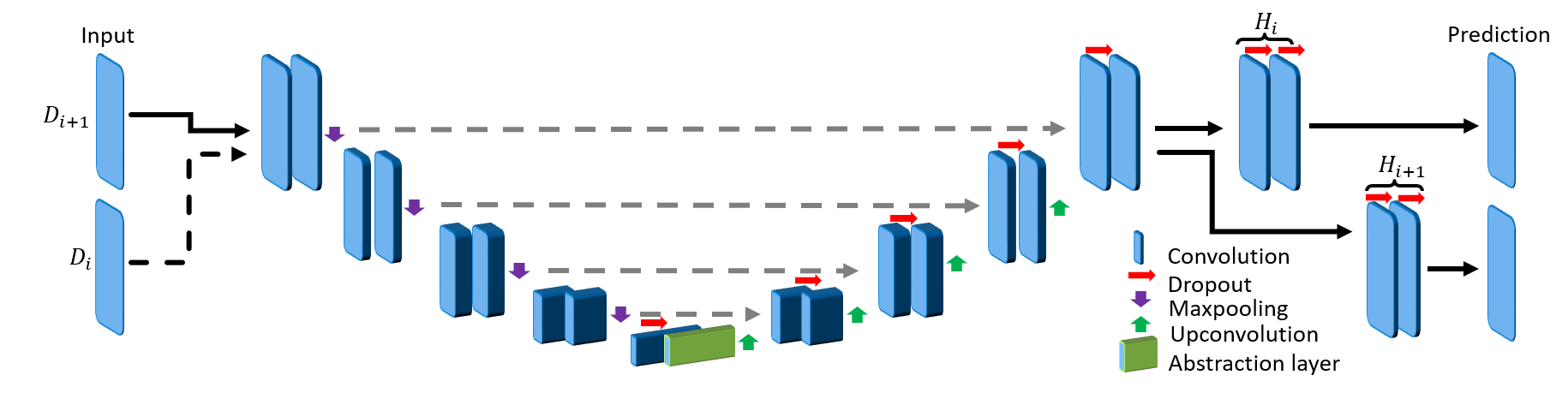}
    \caption{Schematic of the proposed \vR{convolutional network at incremental step $i+1$}. Additional layers (``Head'') at step $i+1$ are shown with $H_{i+1}$. Second layer at coarsest level is called \textit{abstraction layer}. $D_i$ denotes the \emph{exemplar data}\vR{ in Sec.\,\ref{sec:representative}}.}
    \label{fig:schematic}
\end{figure}

\subsection{Conservation of Prior Knowledge with Distillation}

In medical image analysis, a lot of anatomies show similarity in their statistical priors (i.e.,\ bones).
Similarly to the idea of finetuning a pretrained VGG~\cite{Simonyan2014very} network for different digital image classification task, one can train a model with a dataset of anatomy \vR{$A_{x}$} and then finetune it for a 
new dataset of anatomy \vR{$A_{y}$}, as to utilize knowledge obtained from dataset of \vR{$A_{x}$} when learning \vR{$A_{y}$}. 
Based on the application, maintaining segmentation capability of anatomy \vR{$A_{x}$} can be equally important (e.g.\ functional modeling), albeit the two annotated dataset being collected from different patients/studies.
It is possible to use distillation loss~\cite{hinton2015distilling,rebuffi2016icarl} in order to retain the segmentation accuracy of \vR{$A_{x}$}, while exploiting learned prior statistical knowledge to better learn more limited resources of \vR{$A_{y}$}.

Inspired by the idea (LwF~\cite{li2016learning}) from classification, we propose a segmentation framework for class-incremental learning.
At an initial training phase, we train a U-Net~\cite{Ronneberger2015unet} like network (cf.\,Fig.~\ref{fig:schematic} \vR{without $H_{i+1}$}) with a first dataset $D_\mathrm{init}$ using desired classification loss $L_\mathrm{c}$.
In order to disambiguate between initial and incremental training, we will refer to the old and new steps as $i$ and $i+1$, respectively.
Prior to incremental training, all available dataset $D_a = D_\mathrm{i+1} \cup D_\mathrm{i}$ is passed through the pretrained network in order to produce prediction probabilities for the old classes, including their respective background.
For old classes $c_i$, these predictions $p^{c_i}$ are then used for retaining the network's segmentation accuracy.
For incremental training, a new network is then constructed with an additional head $H_{i+1}$ as shown in Fig.~\ref{fig:schematic}.
The parameters for the the network (cf.\,Fig.\,\ref{fig:schematic}) \vR{except for $H_{i+1}$} are loaded from the previously trained model.
During incremental training, for given mini-batch $B$ we optimize the following loss:
\begin{equation}
    L_\mathrm{total}= 
    \begin{cases}
      \alpha L_\mathrm{c} + (1-\alpha) L_\mathrm{d}, & \text{if}\ B \in D_\mathrm{i+1} \\
      L_\mathrm{d}, & \text{otherwise}
    \end{cases}
\end{equation}
where $\alpha \in [0,1]$ is a weighting scalar, and $L_\mathrm{d}$ is the distillation loss defined as:
\begin{equation}
L_\mathrm{d} = \sum_{c_i^{(j)} \in c_\mathrm{i}} p^{c_i^{(j)}}_\mathrm{i}\log(y^{c_i^{(j)}})
\end{equation}
where $p^{c_i^{(j)}}$ and $y^{c_i^{(j)}}$ are the predicted probabilities for class $j \in c_i$ for the initially trained network and the old class heads $H_i$ in the new network.

While it would be ideal to retain all prior training dataset when introducing a new label, this is not a scalable solution due to computational challenges discussed before. 
An extreme case is to remove all prior data from incremental learning; i.e.,\ $D_\mathrm{i} = \{\}$~\cite{li2016learning}.
The method proposed above is called in the following as LwfSeg.
When incrementally learning with datasets of similar properties, LwfSeg may prove sufficient to preserve old class segmentation capacity.
Unfortunately, there can be significant amount of variation in image data across different iterations of training, e.g.\  different MR field inhomogeneity due to different patient profiles, differences across used machine brands.
This can lead to a false guidance of the network with an effort to retain old class knowledge.

\subsection{Selecting Representative Samples}
\label{sec:representative}

One can select a subset of the training data to be kept for future incremental learning processes. 
For classification tasks, a potential approach to choosing representative samples is to observe the feature space right before the final network layer (i.e.,\ embedding space)~\cite{rebuffi2016icarl,hausser2017learning}, where each input sample is represented with a class-discriminating vector. 
 
Although an intuitive extension for segmentation is to use the embedding space, this is not directly applicable.
Finding ``most representative'' pixels per class in the embedding space is not very helpful.
Even if one aggregates a representativeness metric over all pixels at the embedding space as to get a scalar value per input image, it is not clear how to account for the ratio of the foreground pixels accordingly. 
Therefore, we propose the following two alternatives.

\noindent\textbf{Maximum Set Coverage over Most Certain Sample Abstractions:}
Using dropout layers, one can compute a trained models confidence for a given input image $I$ through Monte Carlo estimates~\cite{gal2015dropout} by getting inference $t_{\mathrm{MC}}$ times. 
A typical way to get a scalar uncertainty value from an image is to aggregate the uncertainty over all pixels. 
In~\cite{yang2017suggestive}, in a microscopy segmentation context, authors select samples based on maximum uncertainty for maximizing performance gain in a next round of training.
An ideal exemplar set should contain slices with high confidence (i.e., least uncertainty) for class-incremental task, in order to prevent ``catastrophic forgetting.'' 
For typical medical images where foreground labels are underrepresented, or completely missing in a slice when this anatomy is not intersecting, high confidence samples are likely to fall on images with only background. 
As a remedy, one can choose $k_c$ most \textit{certain} slices for each class among images where corresponding class label exists, creating a ``most'' \textit{certain} set $S_c^j$ for every class $j$.

As an effort to further reduce the kept training data, we aim to select a \textit{representative} subset of $S_c^j$.
Similarly to~\cite{yang2017suggestive}, we use spatially averaged activations at the \textit{abstraction-layer} (c.f. Fig.\ref{fig:schematic}) of our network to represent a given image as a vector $I^{R}_\mathrm{abs}$.
In an iterative fashion, we then create a \textit{representative} set $S_r^j \subseteq S_c^j$ with $k_r$ elements for each class $j$ in order to maximize set coverage~\cite{Hochbaum1997approximating} over the full training set $S_a$, using cosine similarity between each $I^{R}_\mathrm{abs}$.
We call this method Abstraction exemplar-based incremental Segmentation (AeiSeg).
Although one can use a faster method to pick samples for the set $S_r^j$, the iterative set coverage approach ensures maximum set coverage even if in future some of the later picked exemplar samples need to be removed; i.e.,\ storage constraints.

\noindent\textbf{Maximum Set Coverage over Content Distance:}
Albeit being a fast proposition, similarity across spatially averaged activations~\cite{yang2017suggestive} at \textit{abstraction-layer} have questionable image representation capability, as no objective function directly optimizes for \textit{image representation}.
Instead, we use the activations of a pre-trained VGG network at multiple layer levels to describe ``content'' of an input image, which is inspired by~\cite{gatys2016image}.
A VGG network trained on ImageNet~\cite{Russakovsky2014imagenet} has convolutional filters tuned for object recognition and localization task.
Hence, its layer responses aim to distinctively represent objects present in a given image invariant to their spatial location.
Based on this, a distance metric defined over the activations of such a network can give an accurate relative quantification of any two images from a dataset.
Let $R^{l,f}(I_i) \in \mathcal{R}^{w_l, h_l}$ be the activation response of a VGG16 network at layer~$l$, filter~$f$, width $w_l$ and height $h_l$ at the corresponding layer for a given image $I_i$.
We compute the \textit{content distance}
\begin{equation}
	d_\mathrm{cont} = \sum_{l \in L,f \in F_l} (R^{l,f}(I_i) - R^{l,f}(I_j))^2
\end{equation}
as the mean squared distance between activation responses of two images $I_i$ and $I_j$,
where $L$ and $F_l$ are the sets of all convolutional layers and their respective filters of the trained VGG network. 
We call this method Content Representativeness-based incremental Segmentation (CoRiSeg).

\section{Experiments and Results}
\label{sec:Experiments}

\begin{figure}[t]
 
\begin{subfigure}{0.35\textwidth}
\includegraphics[width = \textwidth, trim= 0.7cm 0cm 1.05cm 0cm, clip]{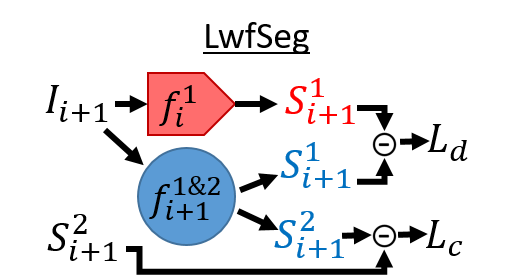}
\label{fig:lwfseg}
\end{subfigure}%
\begin{subfigure}{0.35\textwidth}
\includegraphics[width = \textwidth, trim= 0.7cm 0cm 1.0cm 0cm, clip]{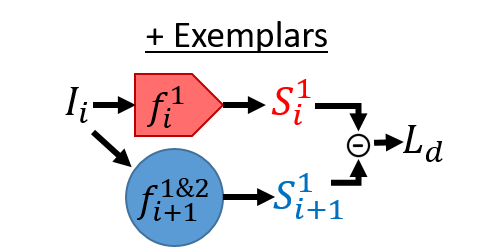}
\label{fig:exemplar}
\end{subfigure}%
\begin{subfigure}{0.295\textwidth}
\centering
    \begin{tabular}{r|r|r}
    \multicolumn{3}{c}{\underline{Experimental Scenarios}} \\
\multicolumn{1}{c|}{} & \multicolumn{1}{c|}{initTrain} & incTrain   \\ \hline
Case 1                & 4 Hum                           & 4 Scap      \\
Case 2                & 6 Scap                          & 1 Hum       \\
Case 3                & 4 Hum                           & 3 Scap*
\end{tabular}
\label{tbl:scenarios}
\end{subfigure}
\caption{Learning an incremental network $f_{i+1}^{1\&2}$ for classes $1\&2$ with new images $I_{i+1}$ and annotations of new structures $S_{i+1}^2$, given a pre-trained and frozen network  $f_i^1$. Left: Representation of LwfSeg. Middle: \vR{Additional loss in AeiSeg and CoRiSeg for augmenting the new network (left)} with exemplar images $I_i$. Right: Experimental scenarios depicting initial (init) and incremental (inc) datasets for humerus (Hum) and scapula (Scap). Case 3 incTrain (*) was conducted on a different MR sequence (water-saturated Dixon).}
\label{fig:incrementalAndTable}
\end{figure}

\begin{figure}[t]
\centering
	\begin{subfigure}[b]{0.332\textwidth}
    \centering
        \includegraphics[width=0.95\textwidth, trim= 0cm 0.0cm 0cm 0cm, clip]{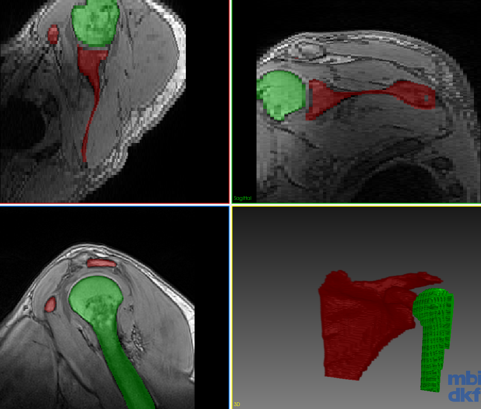}
        \caption{Gold Standard}
        \label{fig:qualRes:GS}
    \end{subfigure}%
    \begin{subfigure}[b]{0.332\textwidth}
    \centering
        \includegraphics[width=0.95\textwidth, trim= 0cm 0.0cm 0cm 0cm, clip]{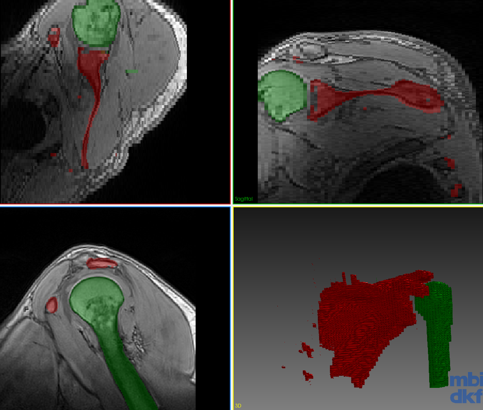}
        \caption{CoRiSeg Case 1}
        \label{fig:qualRes:case1CoRiSeg}
    \end{subfigure}%
    \begin{subfigure}[b]{0.332\textwidth}
    \centering
        \includegraphics[width=0.95\textwidth, trim= 0cm 0.0cm 0cm 0cm, clip]{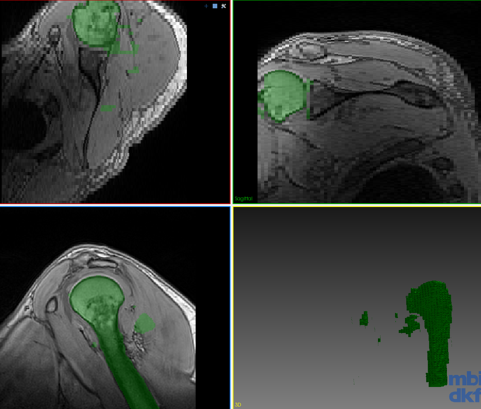}
        \caption{HumSeg Case 2}
        \label{fig:qualRes:case2HumSeg}
    \end{subfigure}%
    \\%
    \begin{subfigure}[b]{0.332\textwidth}
    \centering
        \includegraphics[width=0.95\textwidth, trim= 0cm 0.0cm 0cm 0cm, clip]{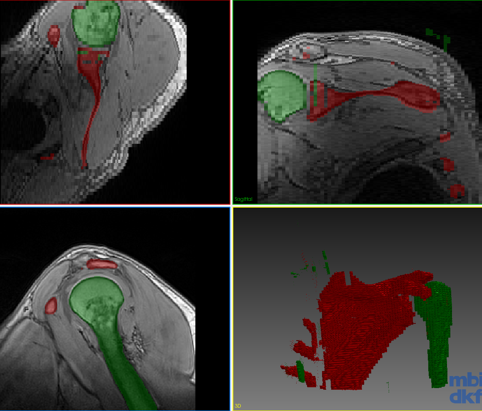}
        \caption{CoRiSeg Case 2}
        \label{fig:qualRes:case2CoRiSeg}
    \end{subfigure}%
    \begin{subfigure}[b]{0.332\textwidth}
    \centering
        \includegraphics[width=0.95\textwidth, trim= 0cm 0.0cm 0cm 0cm, clip]{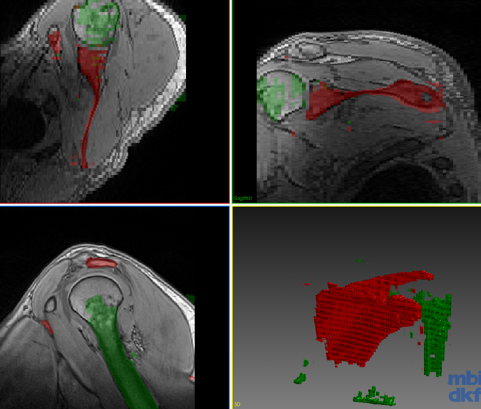}
        \caption{LwfSeg Case 3}
        \label{fig:qualRes:case3LwfSeg}
    \end{subfigure}%
    \begin{subfigure}[b]{0.332\textwidth}
    \centering
        \includegraphics[width=0.95\textwidth, trim= 0cm 0.0cm 0cm 0cm, clip]{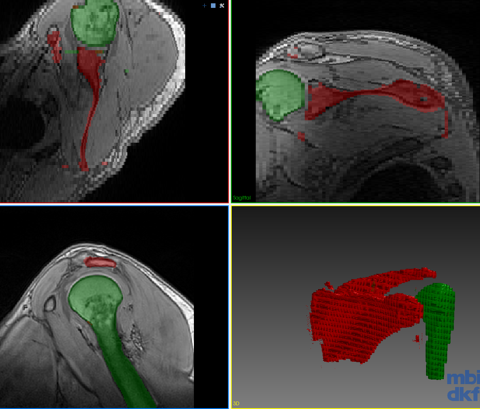}
        \caption{CoRiSeg Case 3}
        \label{fig:qualRes:case3CoRiSeg}
    \end{subfigure}%
\caption{Segmentation of the test volume with different methods.}
    \label{fig:qualRes}
\end{figure}

Our experimental dataset consists of 9 Dixon sequences of left shoulder collected with 1.5 Tesla at resolution of 0.91\,mm x 0.91\,mm x 3\,mm, corresponding to 192x192x64 voxel resolution.
Humerus and scapula bones were annotated by an expert.
Our goal is to combine knowledge from different data for a network that can segment both anatomical structures.
We evaluated our proposed method for the three scenarios shown in the table in Fig.~\ref{fig:incrementalAndTable}.
One volume each was fixed randomly for validation and testing each of all scenarios. 
The first scenario (Case 1) tests a typical setting where different anatomies were of interest and thus annotated in separate studies.
The second scenario (Case 2) aims to observe advantages of incremental training with minimal effort, i.e.,\ incrementally annotated data, giving insight on an extreme case where a single volume annotation is provided.
The last scenario (Case 3) studies the feasibility of combining learned segmentation information from different anatomy and images of different contrast.
The methods were implemented with Tensorflow~\cite{tensorflow2015whitepaper} and ran on an Nvidia Titan X GPU.
\vR{Proposed network is implemented in 2D, hence $64$ image samples per volume given in Fig.~\ref{fig:incrementalAndTable}.}
For a fair comparison, we fixed all parameters across different models to $k_c$\,$=$\,$50$, $k_r$\,$=$\,$30$, $\alpha$\,$=$\,$0.5$, $t_{\mathrm{MC}}$\,$=$\,$29$, batch size of 8 images, and trained all models for 1000 epochs. 
Used network (cf.\,Fig.\ref{fig:schematic}) has a first convolutional layer with 64 filters and the amount of filters double at every coarsening level. 
Each convolutional layer is proceeded with a batch normalization and ReLU activation.
For CoRiSeg, we use a VGG16 network~\cite{Simonyan2014very} pre-trained on ImageNet~\cite{Russakovsky2014imagenet}.
While a VGG trained on a medical image set would be expected to provide more accurate $d_\mathrm{cont}$ score, training set of ImageNet is not matchable by any annotated medical database.
We used Dice similarity coefficient and average symmetric surface distance for evaluating segmentation performance across tested methods (cf.\,Table~\ref{tbl:quantitativeResults}).
We compared our proposed methods: LwfSeg with its extensions with exemplar sets AeiSeg and CoRiSeg.
Upper bound cases are presented with networks trained on only a given anatomy/dataset, i.e.,\ without any incremental learning and hence without the need to preserve ``old'' (extra) information.
We also show results from finetuning for comparison, although catastrophic forgetting is a known problem.
In Fig.~\ref{fig:qualRes}, we showcase qualitative results from different scenarios (cf.\ table in Fig.~\ref{fig:incrementalAndTable}).

\begin{table}[t]
\centering
\caption{Dice coefficient [\%] and average symmetric surface distance (SurfDist) [mm] of networks trained only for \vR{humerus} (HumSeg) and scapula (ScapSeg), finetuning, and our proposed incremental learning methods (cf.\ table in Fig.~\ref{fig:incrementalAndTable}). Best scores of incremental methods are shown in bold.}
\label{tbl:quantitativeResults}
\begin{tabular}{r|rcc|cc|cc|cc|cc|cc}
\multicolumn{2}{c}{}            & \multicolumn{4}{c|}{Case 1}                                                                                 & \multicolumn{4}{c|}{Case 2}                                                                                  & \multicolumn{4}{c}{Case 3}                                                                                                  \\ \cline{2-14} 
\multicolumn{2}{c}{}            & \multicolumn{2}{c|}{Dice}                           & \multicolumn{2}{c|}{SurfDist}                        & \multicolumn{2}{c|}{Dice}                            & \multicolumn{2}{c|}{SurfDist}                        & \multicolumn{2}{c|}{Dice}                                    & \multicolumn{2}{c}{SurfDist}                                 \\ \cline{2-14}
\multicolumn{1}{c}{} & \multicolumn{1}{c|}{Method}     & hum                     & scap                      & hum                       & scap                      & scap                     & hum                       & scap                      & hum                       & hum                     & scap                               & hum                       & scap                              \\ \hline
\multirow{2}{*}{\shortstack[l]{\scriptsize Upper-\\\scriptsize bound}} & \multicolumn{1}{r|}{HumSeg}     & 96.7                    & -                         & 0.36                      & -                         &                          & 82.9                      & -                         & 5.38                      & 96.7                    & -                                  & 0.36                      & -                                 \\
& \multicolumn{1}{r|}{ScapSeg}    & -                       & 88.1                      & -                         & 0.67                      & 88.5                     & -                         & 0.66                      & -                         & -                       & 84.6                               & -                         & 0.93                              \\
\hline
{\scriptsize Baseline} & \multicolumn{1}{r|}{finetuned} & \multicolumn{1}{l}{0.1} & \multicolumn{1}{l|}{86.2} & \multicolumn{1}{l}{41.22} & \multicolumn{1}{l|}{1.08} & \multicolumn{1}{l}{13.4} & \multicolumn{1}{l|}{79.8} & \multicolumn{1}{l}{16.41} & \multicolumn{1}{l|}{7.54} & \multicolumn{1}{l}{0.0} & \multicolumn{1}{l|}{\textbf{83.8}} & \multicolumn{1}{l}{65.93} & \multicolumn{1}{l}{\textbf{1.21}} \\ \hline
\multirow{3}{*}{\scriptsize Proposed} & \multicolumn{1}{r|}{LwfSeg}     & 95.9           & \textbf{87.5}             & \textbf{0.88}             & \textbf{1.51}             & 73.2                     & 87.3                      & 6.08                      & \textbf{7.12}             & 66.0                    & 79.1                               & 13.32                     & 2.86                              \\
& \multicolumn{1}{r|}{AeiSeg}     & 96.1           & 82.9                      & 1.31                      & 2.27                      & 74.8                     & 63.5                      & 7.09                      & 36.52                     & 94.2                    & 76.7                               & 2.68                      & 3.20                              \\
& \multicolumn{1}{r|}{CoRiSeg}    & \textbf{96.3}           & 82.8                      & 0.89                      & 2.30                      & \textbf{78.7}            & \textbf{90.2}             & \textbf{5.38}             & 13.51                     & \textbf{94.8}           & 78.7                               & \textbf{1.56}             & 2.03                             
\end{tabular}
\end{table}

\section{Discussions}
\label{sec:discussions}

As seen, with finetuning, the shared network body gets re-tuned to adapt to the new incremental data, almost completely forgetting the initial classes. 
Proposed segmentation extension of learning without forgetting (LwfSeg) performs relatively well in all cases.
For every scenario, both proposed methods using exemplar sets either outperform or achieve performance as close to LwfSeg for the old class.  
CoRiSeg achieves the highest Dice score for the old data, suggesting that for selecting exemplars the maximum set coverage over content distance is more effective than averaging at abstraction-layer (AeiSeg).
In addition, in Case 2, where the incremental dataset is severely handicapped, both LwfSeg and CoRiSeg surprisingly outperform HumSeg.
While it is expected for a network trained on 1 volume (64 images) to perform poorly, incremental networks are seen to achieve higher segmentation performance, suggesting that shared-body layers potentially learned to extract \textit{bone-generic knowledge}.
Should this be shown for a wider range of bone structures, it would be critically relevant for orthopedic applications in the future.
When the incremental dataset is introduced from a different imaging sequence in Case 3, one can see the great advantage of keeping exemplar samples; i.e., 28.8\% increase in Dice score of CoRiSeg compared to LwfSeg. 
While the performance difference is less obvious for the new class, the change in old class scores suggests distillation loss to have provided false ``guidance'' on the new dataset with LwfSeg, i.e.\ trying to retain old class segmentation performance without any exemplar samples.
Since finetuning does not need to remember the appearance of humerus (bone) in the other image modality, it outperforms with scapula in Case\,3.
\vR{We expected VGG trained for object classification (on ImageNet) to select better exemplar images for our task.
Indeed, compared to CoRiSeg, using the UNet trained by us for $d_{\mathrm{cont}}$ yielded 1.6\%, 0.7\%, and 0.07\% worse Dice, respectively, for each Case.}

Note the high average symmetric surface distance in some of the proposed incremental methods, i.e., AeiSeg and CoRiSeg in Case 2; LwfSeg in Case 3.
These are due to small blobs of false positives far from the target anatomy (cf.\,Fig.\,\ref{fig:qualRes}).
These blobs could possibly be removed with a trivial post processing step (e.g. morphological operations, largest connected region, conditional random fields, user input), which is beyond the objective of this paper.
\vR{Additional randomized hold-out test sets for Case 3 showed little variation ($\approx$2\% Dice) in results, while the proposed AeiSeg and CoRiSeg were still over 27\% Dice better than LwfSeg in retaining old class info.
We will conduct extensive evaluations in future.} 

\section{Conclusions}
\label{sec:conclusion} 

In this work, we have proposed a solution for applying class-incremental learning to \textit{segmentation} using a distillation objective function (LwfSeg). 
However, with increasing size of labels and variability in medical images, we have shown that LwfSeg may become suboptimal without an exemplar dataset.
To address this, we have proposed two novel methods to select \textit{representative} images based on abstraction layer response (AeiSeg) and content distance (CoRiSeg)\vR{; which imposes no restrictions on the incremental data size or \#classes}.
We have evaluated the proposed frameworks on three different scenarios that often exist in medical image analysis community and shown that the proposed methods achieve performance similar to the upper-bound conditions.
LwfSeg showed promising efficiency in retaining old class segmentation performance, which was improved further with proposed extensions with exemplar selections.
To the best of our knowledge, this work is the first to show class-incremental learning in medical image segmentation (LwfSeg), and its extensions with intelligent exemplar selection (AeiSeg \& CoRiSeg), CoRiSeg being our favorite with its intuitive design and higher performance.
In the future, we will extend incremental training to additional anatomical structures and imaging modalities.

\noindent\textbf{Acknowledgements.}
Funded provided by the Swiss National Science Foundation (SNSF) and a Highly-Specialized Medicine grant of the Canton of Zurich.

\bibliographystyle{splncs03}
\bibliography{main}

\end{document}